\documentclass{svproc}
\usepackage{url}

\usepackage{graphicx}
\usepackage{float}
\usepackage{adjustbox}
\usepackage{listings}
\usepackage[table]{xcolor} 
\usepackage{tabularx}
\usepackage{hyperref}
\usepackage{threeparttable}

\newcommand{\ours}{\textsc{AIR}}
\usepackage{pgfplots}
\pgfplotsset{compat=1.18}
\usepackage{graphicx} 
\usepackage{adjustbox}
\usepackage{listings}
\usepackage[table]{xcolor} 
\usepackage{svg}
\usepackage{tabularx}
\usepackage{hyperref}

\usepackage{booktabs}
\usepackage{siunitx}         
\PassOptionsToPackage{hyphens}{url}
\usepackage{url}
\usepackage{hyperxmp}
\usepackage{authblk}

\newcommand{\DTG}{\mathcal{G}_{\mathrm{DTG}}}
\newcommand{\R}{\mathcal{R}}

\definecolor{paleblue}{RGB}{235,245,255} 
\definecolor{codegreen}{rgb}{0,0.6,0}
\definecolor{codegray}{rgb}{0.5,0.5,0.5}
\definecolor{codepurple}{rgb}{0.58,0,0.82}
\definecolor{backcolour}{rgb}{0.97,0.97,0.97}

\lstdefinestyle{mystyle}{
    backgroundcolor=\color{backcolour},   
    commentstyle=\color{codegreen},
    keywordstyle=\color{magenta},
    numberstyle=\tiny\color{codegray},
    stringstyle=\color{codepurple},
    basicstyle=\ttfamily\footnotesize,
    breakatwhitespace=false,         
    breaklines=true,
    columns=fullflexible,
    captionpos=b,                    
    keepspaces=true,                 
    numbers=left,                    
    showspaces=false,                
    showstringspaces=false,
    showtabs=false,                  
    tabsize=4
}

\begin{document}
\mainmatter              
\title{Autonomous Issue Resolver: \\Towards Zero-Touch Code Maintenance}
%
%
\author{Aliaksei Kaliutau}
%
%
%
\institute{Imperial College London, London, UK,\\
\email{aliaksei.kaliutau24@imperial.ac.uk},\\ WWW home page:
\texttt{https://akaliutau.github.io/}
}

\maketitle              

\begin{abstract}
Recent advances in Large Language Models have revolutionized function-level code generation; however, repository-scale automated program repair remains difficult because the cause of a defect and its observable failure are often distributed across files, layers, and abstractions, while LLM agents operate under limited context. This paper presents Autonomous Issue Resolver (AIR), a multi-agent repair system centered on a Data-First Transformation Graph (DTG), which represents data states as nodes and transformations as edges. The representation enables the agent to localize faults by following the data line rather than browsing files or relying only on lexical retrieval. AIR combines DTG navigation, targeted editing, and test-driven validation under a risk-aware control policy. The system is evaluated on SWE-bench Verified/Lite and on an exploratory set of 100 issues from 10 open-source Python repositories. AIR achieves 87.1\% on SWE-bench Verified and 73.5\% on SWE-bench Lite; in the OSS study it resolves 70/100 issues under a fixed tool and a time budget. The ablation results show that the removal of RL reduces the resolution from 87.1\% to 82.5\%, while the removal of DTG reduces it to 56.2\%, indicating that data-first repository representation is the main performance driver. These results suggest that explicit data-lineage navigation is a promising alternative to file-first retrieval for autonomous software maintenance.

\keywords{Automated Program Repair, repository-scale reasoning, data transformation graph, LLM agents, Data Lineage}
\end{abstract}
\section{Introduction}

In the 21st century, software engineering has evolved from a niche discipline into the digital backbone of the global economy, generating trillions in value and employing hundreds of millions of people worldwide. Today, software is the invisible infrastructure that connects commerce, communication, healthcare, transportation, and nearly every facet of modern life.

Software maintenance dominates engineering effort: debugging and maintenance account for 50--70\% of development resources in many organizations~\cite{cisq2020}. However, repository-scale automated program repair remains difficult because the cause of a failure and its manifestation are often distributed across files, layers, and abstractions. Current LLM-based coding agents can generate plausible local patches, but remain less reliable when the task requires repository-level fault localization, constrained editing, and regression-aware validation.

The central challenge is not code generation in isolation, but \emph{systemic reasoning}: mapping an issue report to a minimal, safe set of edits under repository-wide constraints such as APIs, invariants, tests, and deployment rules. This requires both a repository representation that preserves causal structure beyond lexical similarity and a control policy that decides what to inspect, where to edit, and when to stop.

Modern models can synthesize non-trivial patches and reason about code structure, yet this progress has not translated into robust performance on realistic repository-level tasks: on benchmarks such as SWE-bench~\cite{jimenez2024swebench}, resolution rates remain modest, especially for open-source models.

This paper refers to this gap as the \emph{context crisis} of repository-scale APR. At small scales, LLMs can receive all relevant code in a single prompt. At repository scale, however, the model must operate under strict context limits while the bug's cause and manifestation are separated across files, layers, and abstractions. Existing tools address this by improving retrieval and by adding agentic planning, but they leave the underlying representation of the repository largely unchanged: a directory of files, occasionally augmented with control-centric graphs.

This work argues that this file-first, control-centered view is misaligned with the causal structure of many real-world bugs and how the development is done in the real world. Human developers rarely debug or write new code by reading files. Instead, they trace the \emph{flow of data}, following a specific object or variable as it is transformed and routed through the system. This paper proposes making the data-first mental model explicit and turning it into the primary substrate on which a LLM-based repair agent lives.

The proposed approach is evaluated on SWE-bench Verified/Lite and an exploratory OSS study; the main results are reported in Section~\ref{sec:results}.

The main contributions of this work are as follows.

\begin{itemize}
\item A \emph{Data-First Transformation Graph} (DTG) that models data states as nodes and transformations as edges, enabling causal fault localization beyond file-first retrieval.
\item AIR, a multi-agent repair architecture that combines DTG navigation, targeted editing, and test-driven validation under a risk-aware control policy.
\item An empirical evaluation on the SWE-bench and an exploratory OSS repository study, with isolation of the effects of DTG and the control policy.
\end{itemize}

The remainder of the paper first summarizes the limits of current coding assistants and APR systems, then positions AIR against related work, introduces the DTG representation and AIR architecture, and finally evaluates the approach on public benchmarks and an exploratory repository-level OSS study, followed by explanations, limitations, and future work.

\section{Related Work}

\subsection{Background and Motivation}

AI-assisted software development is already well established in practice. Tools such as GitHub Copilot~\footnote{\url{https://docs.github.com/copilot}}, Amazon CodeWhisperer~\footnote{\url{https://docs.aws.amazon.com/codewhisperer/}}, Tabnine~\footnote{\url{https://docs.tabnine.com/main}}, Claude~\footnote{\url{https://claude.ai}}, and chat-based coding assistants are widely used; however, most of them still optimize developer assistance rather than autonomous repository-scale maintenance (see also Appendix, Table~\ref{tab:landscape}).
Most existing products are iterating on developer help.

Table~\ref{tab:landscape} separates \emph{professional}, \emph{basic}, and \emph{specialized} offerings. Professional tools optimize \emph{ergonomics} for humans-in-the-loop (IDE integration, fast completions), but remain fundamentally \emph{reactive} and local in scope. Basic tools accelerate \emph{scaffolding} and prototyping, yet rarely manage repository-scale dependencies or change risk. Specialized systems reach \emph{SOTA} in narrow settings (e.g., competitive programming or algorithm discovery), but are not designed for end-to-end maintenance workflows.

\begin{table}[t]
\centering
\small
\caption{Capability taxonomy: where existing tools help - and where they fall short for autonomous maintenance. We classify existing tools by their primary interaction model and identify the specific limitations that prevent them from achieving zero-touch maintenance.}
\label{tab:capabilities}
\begin{tabular}{@{}p{2.4cm}p{4.4cm}p{5.4cm}@{}}
\toprule
\textbf{Category} & \textbf{Typical Features} & \textbf{Limits for Maintenance} \\
\midrule
Completion/ Suggestion & Next-line blocks, boilerplate, refactors & Local context only; no global impact modeling; no gating against risk. \\
Conversational Dev Assist & Explain, edit, diagnose from chat & Quality depends on prompt curation; inconsistent across large repos. \\
Repo Search\& Retrieval & Embedding search, code Q\&A, cross-ref & Weak semantics beyond lexical/embeddings; limited data-/control-flow. \\
Automated Program Repair (APR) & Patch synthesis for unit-failing bugs & Often single-hunk fixes; brittle across multi-file causal chains. \\
Test Generation & Unit/integration test synthesis & Improves coverage, but does not plan changes or manage regressions. \\
Agentic Tool Use & Editors, terminals, CI tools via agents & Without risk-aware policy, tends to oscillate or over-edit. \\
\bottomrule
\end{tabular}
\end{table}

\subsection{Limitations of existing tools}

Current LLM-based agents, such as SWE-Agent and Claude typically interact with repositories through a file-system interface: they list directories, open files, and issue search commands. Each action consumes context and forces the model to maintain a growing internal representation of control and data dependencies. As repositories grow, this misalignment becomes the main bottleneck: the model "thinks" at the level of dataflow but sees the world as files and control structures. 

Retrieval-Augmented Generation (RAG) is the standard strategy for scaling LLMs to large corpora. In code domains, RAG usually combines a lexical search with dense embeddings of code snippets. For task prompts involving a particular entity (e.g., fixing some issue that involves a \texttt{User} class), the system retrieves semantically similar snippets for the model to condition on.

However, semantic similarity is not the same as \emph{causal} relevance:

\begin{itemize}
\item If a bug involves corrupt user IDs, a vector search for \texttt{User} will fetch class definitions, UI components, and schema migrations but may miss a generic sanitizer function that silently strips or hashes IDs without ever mentioning the word "user".
\item In data pipelines, failures may be caused by a shared transformation used in multiple domains. Semantically similar code snippets may not share a causal influence on the failing data.
\end{itemize}

This is increasingly common in modern codebases that use loosely coupled design patterns, where relevant logic may reside in semantically disconnected function clusters that are not called directly (e.g. aspect-oriented programming, dependency injection, and so on). 

Traditional RAG treats the code as an unordered bag of textual fragments. Even graph-based retrieval (e.g., GraphRAG) typically uses graph structure only to improve retrieval quality, not as the state space in which the agent itself moves. The agent still \emph{lives} in the file system, reading linearized text.

Taken together, prior work improves code generation, retrieval, or tool use, but the dominant interaction model remains file-first: graphs are typically used to improve retrieval, while the agent still plans and edits in a text/file workspace. AIR differs in two respects. First, the DTG is the agent's primary navigation space, rather than an auxiliary retrieval index. Second, the control policy is tied to validation outcomes, with the goal of reducing over-editing and premature patch acceptance. This positions AIR as a repository-scale repair system focused on causal localization and change risk, rather than as a general-purpose coding assistant.

\subsection{Literature Review}

Transformer-based models trained on large code corpora have substantially advanced software engineering. Early systems such as Codex~\cite{chen2021codex} and AlphaCode~\cite{li2022alphacode} showed that models can synthesize functional code from natural-language descriptions. These advances underpin modern interactive coding assistants. However, while such systems are effective for local "next-block" completion~\cite{yetistiren2023evaluatingcodequalityaiassisted}, they still rely on the human developer to manage repository-wide context and verify correctness.

Agentic frameworks such as ReAct \cite{yao2023react} and Reflexion \cite{shinn2023reflexion},  allowed LLMs to plan, execute tools, and observe outputs. State-of-the-art systems like SWE-agent \cite{yang2024sweagent} and OpenHands \cite{openhands2024} utilize these patterns to interact with repositories via shell commands and file editors, and currently often used as solid baselines. Similarly, AutoCodeRover \cite{zhang2024autocoderover} employs abstract syntax tree (AST) analysis to improve fault localization.

However, these agents largely treat the repository as a file system to be searched and read, a file-first perspective that becomes inefficient when defects span multiple abstraction layers. Evaluations on SWE-bench indicate that such agents can solve isolated tasks but remain weaker on long-horizon repository-level reasoning~\cite{jimenez2024swebench,llmaprsurvey2025}. More recent localization-oriented systems, such as LocAgent~\cite{chen2025locagent}, improve retrieval and fault localization, yet they still operate primarily on text-based or partially structured representations rather than on explicit data lineage.

Since the search over the codebase is essentially a search over unstructured data, it was natural to employ the graph-based methods for localization of relevant snippets of code. The Code Property Graph (CPG) \cite{yamaguchi2014cpg} unified ASTs, control-flow graphs (CFGs), and program dependence graphs (PDGs) into a single structure, primarily for vulnerability discovery. 

Recent efforts have attempted to integrate such graphs with LLMs to improve repository understanding. CodexGraph \cite{liu2024codexgraph} and RepoGraph \cite{ouyang2025repograph} utilize graph databases to enhance Retrieval-Augmented Generation (RAG) by capturing dependencies between files and functions. The Code Graph Model (CGM) \cite{tao2025codegraphmodelcgm} similarly integrates graph structures to aid in logical reasoning. However, these approaches often retain the representation bloat of standard CPGs or focus on improving retrieval recall rather than navigation. Unlike these control-centric models, the proposed approach utilizes a Data-First Transformation Graph (DTG), which compresses the topology to focus on data states and transformations and is aligned with flow-based debugging and optimization practice.

Although most current agents rely on prompt engineering and in-context learning, Reinforcement Learning (RL) has shown promise in discovering novel strategies in complex domains. DeepMind's work on AlphaZero~\cite{silver2017masteringchessshogiselfplay} and AlphaDev~\cite{mankowitz2023alphadev} demonstrated that RL agents can discover algorithms that outperform human-written implementations. In the domain of program repair, early learning-based systems such as Getafix~\cite{bader2019getafix} mined fix patterns from static analysis. AIR incorporates an RL-driven control policy to navigate the DTG, allowing the system to learn stopping conditions and navigation strategies that minimize regression risks. A simple DQN-based online learning scheme is used; RL is not a core dependency or main focus of the proposed method and could be replaced for example by simpler Monte Carlo Tree Search-style graph traversal.

\section{Methodology}

\subsection{A data-first graph representation}

\begin{figure}[!t]
\centering
\includegraphics[width=4.8in]{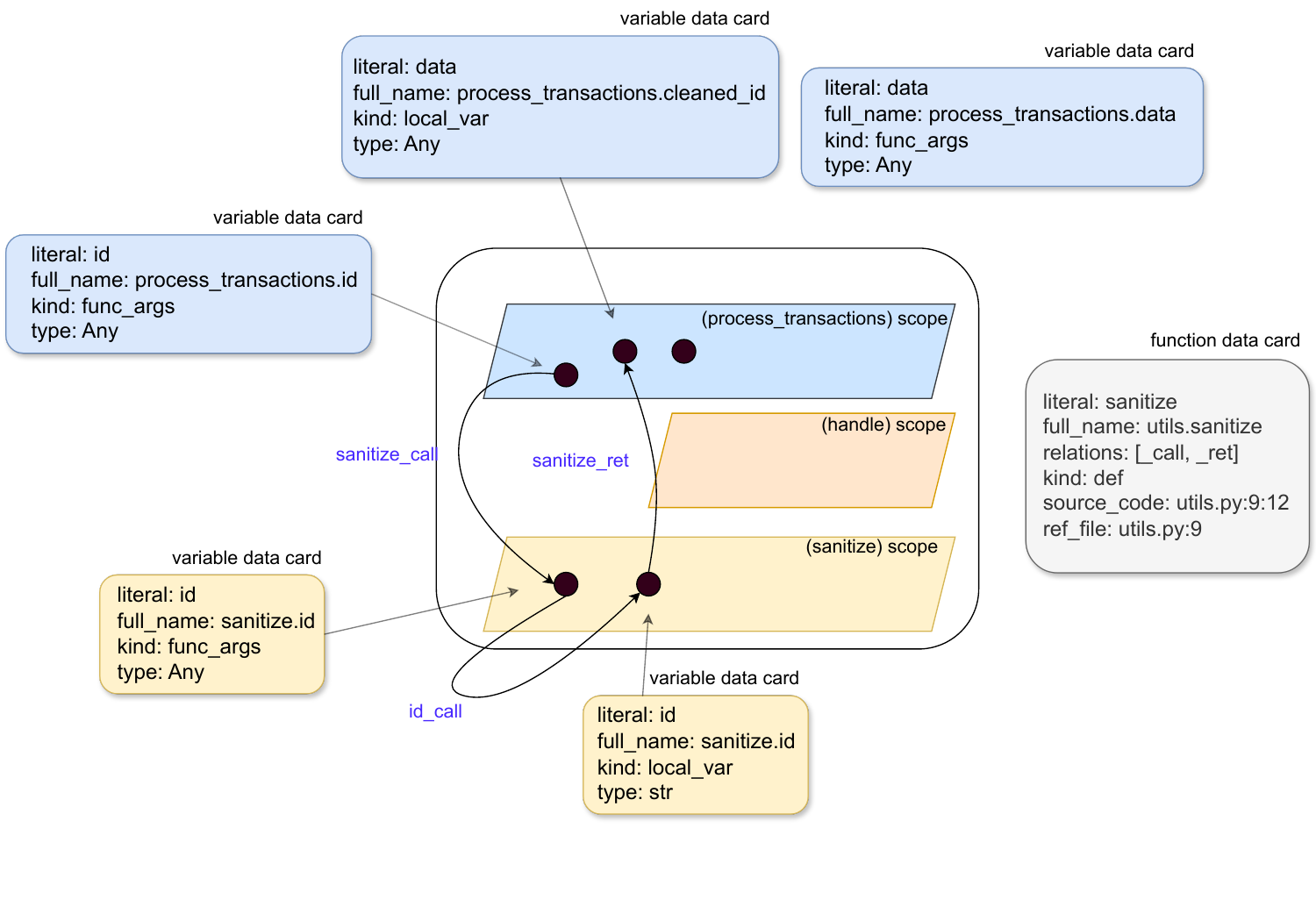}
\caption{The partial visualization of the multi-dimensional Data Transformation Graph for a snippet of code. Nodes contains full description of processed data. Edges are effectively transformation functions (\it{sanitize}, \it{handle}) }
\label{fig:dfg_1}
\end{figure}

This paper proposes a fundamental re-architecture of the agent's environment. Rather than representing a repository as text files, or even as a multi-layer code property graph in which functions and statements are the primary nodes, we focus directly on the \emph{data} that flows through the system.

The \emph{Data-First Transformation Graph} (DTG) is defined as a directed multigraph with:

\begin{itemize}
\item \textbf{Nodes as data states:} Each node corresponds to a semantically meaningful data artifact (e.g., a variable, object, etc) at a specific point in its lifecycle, annotated with type, schema, constraints, and source locations.
\item \textbf{Edges as transformations:} Each edge corresponds to an operator, function, or API call that transforms one data state into another, potentially guarded by control-flow conditions.
\end{itemize}

To illustrate this concept, consider the following snippet of code that implements a toy transaction handle. Figure~\ref{fig:dfg_1} depict a simplified DTG for this code.

\begin{lstlisting}[language=Python]
utils.py
def process_transactions(id, data):
    cleaned_id = sanitize(id)
    return handle(cleaned_id, data)
    
def handle(cleaned_id, data):
    print(f"processed {cleaned_id}")
    
def sanitize(id: Any) -> str:
    id = str(int(id))
    assert len(id) == 4, f"id: wrong length: {id}"
    return id
\end{lstlisting}

This representation compresses away syntactic detail and control-flow noise, allowing the agent to traverse the "causal path" of a variable, effectively pruning irrelevant control flow and syntax noise that typically confuses. Conceptually, the repository ceases to be a library of documents and becomes a navigable network of data lineages.

\subsection{Role of the control policy}

In AIR, reinforcement learning is used to learn \emph{navigation and stopping decisions}, not to generate code tokens directly. Given the current DTG neighborhood, the policy selects actions such as navigating upstream or downstream, inspecting a transformation, invoking validation, or terminating the search. The rewards are tied to successful issue resolution, regression avoidance, and editing efficiency. This makes the control policy risk-aware: trajectories that produce broad or unstable patches are penalized relative to minimal validated fixes.

The second key ingredient is the integration of RL patterns. The design draws inspiration from the successes of systems like DeepMind's AlphaGo and AlphaCode~\cite{silver2017masteringchessshogiselfplay},~\cite{alphacode2022}, which demonstrated how machines can achieve superhuman performance in complex domains not just by mimicking data, but by learning optimal strategies through interaction and self-correction.

Unlike supervised learning (training on fixed code examples), RL allows the AI agent to learn by doing. It generates code, tests it (using automated test suites), observes the results (pass/fail, performance metrics, new bugs introduced) and receives feedback (rewards or penalties). This iterative loop enables the system to find correct and efficient code implementations that might not exist in any training dataset, to optimize code for specific goals, and to self-correct and self-improve. The latter is very important, because it effectively removes the slow human factor (which is the main bottleneck) and allows the system to learn exponentially. The RL loop is implemented in a quite straightforward way, via anchoring objective function to the unit tests

\subsection{Why structured repository reasoning is necessary}

A learned control policy alone does not solve repository-scale repair because the search space remains combinatorial. AIR therefore couples the policy with DTG-based context. The graph narrows exploration to causally relevant data lineages, while the policy determines which neighborhoods to inspect and when enough evidence has been accumulated to move from localization to editing and validation. This division of labor is what makes the approach scalable at repository level.

\section{Algorithm details}

The proposed data-first perspective sits at the intersection of several research lines: code property graphs and static analysis, data-centric optimization and flow-based programming, and graph-based + agentic approaches to LLM-driven software engineering.

\subsection{Code property graphs and representation bloat}

The Code Property Graph (CPG), popularized by tools such as Joern~\cite{joern}, unifies three classic program representations in a single multi-graph: abstract syntax trees (ASTs), control-flow graphs (CFGs), and program dependence graphs (PDGs). In principle, CPGs contain all information needed for both security analysis and program repair.

In practice, however, raw CPGs suffer from extreme representation bloat. The DTG can be viewed as a \emph{semantic compression} of such graphs. By abstracting away AST and CFG nodes and focusing only on data states and the transformations between them, it adopts a "functions-as-edges" topology, in contrast to the "functions-as-nodes" topology of call graphs and many CPG formulations. This aligns the representation with the causal structure of dataflow, rather than the syntactic structure of the code.

\subsection{Data-centric optimization and flow-based programming}

In high-performance computing and compiler optimization, there has been a shift from control-centric intermediate representations (such as linearized IR or basic-block graphs) to data-centric views that emphasize movement and locality. Data-centric frameworks, such as Stateful Dataflow Multigraphs, treat computations as transformations on data streams and have demonstrated significant benefits in optimization and reasoning.

Similarly, flow-based programming (FBP) and ETL (Extract--Transform--Load) pipelines in data engineering model systems as graphs of black-box components connected by typed data channels. While this style is common in specialized domains, it has rarely been applied to general-purpose, object-oriented software in the context of defect localization and repair.

Our DTG formulation can be seen as importing these data-centric ideas into APR. The repository is treated as a large, heterogeneous dataflow graph in which "nodes" are not operators but states, and "edges" encapsulate black-box transformations. 

\subsection{Agentic navigation on graphs}

Context Agent treats the DTG as the world itself. The agent's state is a node (or small subgraph), its primitive actions are graph traversals and inspections, and its perception is restricted to a local neighborhood. Moving the agent "body" into the graph opens up new strategies for search, planning, and exploration based on graph theory and dataflow, rather than ad-hoc file navigation.

\subsection{The Data-First Transformation Graph}
\label{sec:dtg}

To formalize the Data-First Transformation Graph (DTG) we introduce first some definitions.

\subsubsection{Formal definition}

Let a software repository $\R$ be represented as a directed multigraph
$$
\DTG = (V, E),
$$
where $V$ is a set of vertices (data states - see below) and $E$ is a set of edges (transformations).

Each vertex $v \in V$ represents a data state: a specific variable, object, or data artifact at a particular semantic stage. Compared with classical data-flow graphs, DTG nodes are enriched with semantic metadata, not merely variable names.

Each vertex $v$ carries:

\begin{itemize}
\item \textbf{name:} A unique identifier, e.g., \texttt{module.func.var}.
\item \textbf{kind:} Class of data, e.g. is it the argument of function, a global variable, or constant, etc.
\item \textbf{type:} Its static or inferred language-level type, for example, \texttt{List[int]} or \texttt{torch.Tensor}.
\item \textbf{schema:} For structured objects, a description of fields and their types (e.g., {\texttt{\{id:int, name:str\}}}).
\item \textbf{constraints:} Symbolic constraints known at this point, such as shape information or invariants (e.g., \texttt{shape = (3, 224, 224)}).
\item \textbf{metadata:} Source-level metadata, including file path, function, and line span.
\end{itemize}

This information can be partially obtained from static analysis (e.g., SSA form, type inference) and partially hypothesized or refined by the LLM during exploration.

\subsubsection{Edge set: transformations}

Each edge $e_{i \to j} \in E$ represents a transformation from data state $v_i$ to data state $v_j$. An edge exists when there is a concrete operation in the repository that takes the value at $v_i$ and produces the value at $v_j$.

Edges are annotated with:

\begin{itemize}

\item \textbf{plane:} A reference to the view plane that represents on some level of abstraction the transformation, e.g., a function call, and so on.
\item \textbf{source, ref\_file:} Reference points to definition and points of invocation, respectively.
\item \textbf{semantics:} An optional natural-language summary of the transformation's intent, generated and iteratively refined by the LLM (e.g., "increments a counter", "normalizes a tensor").
\end{itemize}

\section{Graph construction pipeline}

Constructing the DTG is the most technically demanding phase. The current implementation uses a "lazy-loading" pipeline that combines static and dynamic information.

\paragraph{Core parser: Tree-sitter and ast:}
Tree-sitter~\cite{treesitter} provides incremental, multi-language AST parsing for languages such as Python, Java, JavaScript, and C++. It is used to:

\begin{itemize}
\item Extract variable declarations, assignments, and function definitions,
\item Identify function calls and their arguments,
\item Recover control-flow constructs for guard extraction.
\end{itemize}

Custom Tree-sitter queries (in \texttt{.scm} files) define language-specific patterns for these elements~\cite{treesitter}, but the downstream DTG schema is language-agnostic. Rather than directly exposing the full internal graph of these tools to the LLM, we project only the data-state and transformation information needed for DTG navigation. A \texttt{neo4j} graph database is used as persistent storage and efficient indexing over node and edge attributes.

\subsection{Agent environment}

\begin{figure}[!t]
\centering
\includegraphics[width=4.8in]{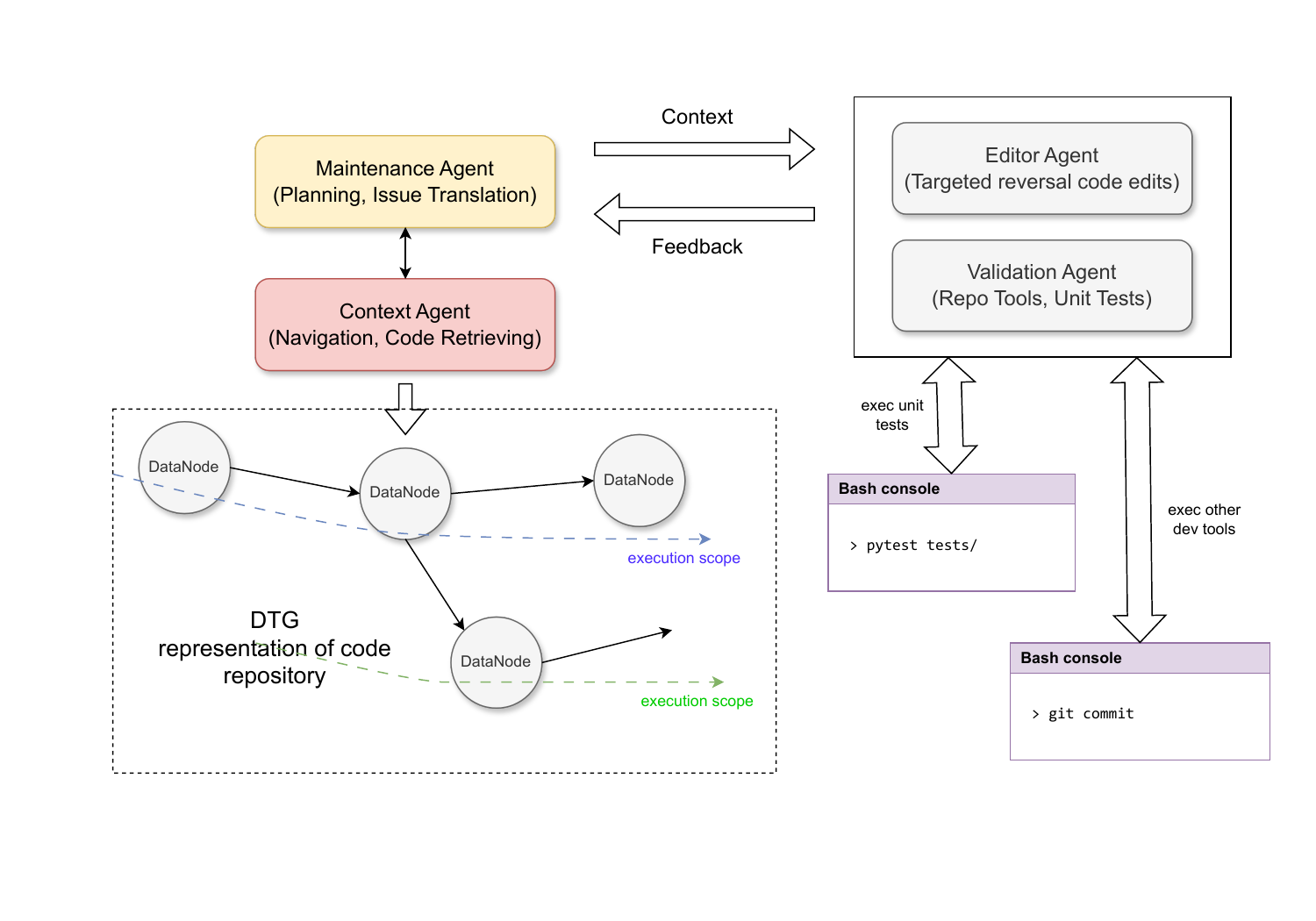}
\label{fig:dataflow}
\caption{The AIR Multi-Agent Architecture. The system employs a decoupled "Plan-Navigate-Execute" loop to manage context window limitations. The Context Agent (pink) operates exclusively within the graph domain to localize faults without reading file content. Once the fault subgraph is identified, the Maintenance Agent (yellow) formulates a high-level repair plan, which is translated into concrete syntax edits by the Editor Agent (gray). This separation prevents the "Semantic Trap" by ensuring the planner is conditioned only on causally relevant data lineages rather than raw repository noise.}
\end{figure}

Agents are modeled with an internal state represented as a structured view over the DTG.

\paragraph{State schema.} A minimal typed state can be defined as:

\begin{lstlisting}[language=Python]
class AgentState(TypedDict):
  current_node_id: str
  visited_nodes: List[str]
  hypothesis: str
  graph_view: SubGraph  # local neighborhood
\end{lstlisting}

The \texttt{graph\_view} restricts the visible graph to a bounded-radius neighborhood around the current node, which is passed to the LLM at each step.

\paragraph{Tools (action space).} The agent is equipped with a small set of graph-oriented tools (Table~\ref{tab:functions}). These tools abstract away file-system operations and present the agent with a clean graph navigation API.

\begin{table}[h]
    \centering
    \renewcommand{\arraystretch}{1.4}
    \begin{tabular}{|p{0.4\textwidth}|p{0.55\textwidth}|}
        \hline
        \textbf{Function Signature} & \textbf{Description} \\
        \hline
        \texttt{navigate(direction= "upstream"|"downstream")} & 
        Move along incoming or outgoing edges from the current node, updating \texttt{current\_node\_id}. \\
        \hline
        \texttt{inspect\_data()} & 
        Return metadata (type, schema, constraints, context) for the current node. \\
        \hline
        \texttt{read\_transformation (target\_node\_id)} & 
        Return code associated with the edge from the current node to \texttt{target\_node\_id}. \\
        \hline
        \texttt{run\_test(test\_id)} & 
        Execute a test case, optionally with dynamic taint tracking, and overlay the resulting execution trace onto the DTG. \\
        \hline
    \end{tabular}
    \vspace{0.1in}
    \caption{Available Navigation and Testing Functions}
    \label{tab:functions}
\end{table}

\section{Experiments and Benchmarks}

Evaluating the performance of AI systems designed for complex, autonomous tasks (such as software development) presents unique challenges. Unlike traditional software, where deterministic output is expected, AI systems often exhibit variability and rely on complex internal behavior. 

\subsection{Benchmarks and Metrics}

The evaluation uses two public benchmark families and one exploratory repository-level study.

\textbf{SWE-bench.} The system is evaluated on SWE-bench Verified and SWE-bench Lite using \emph{\%Resolved with repository tests} as the primary metric~\cite{swebenchsite,swebenchverified}.

\textbf{APR.} HumanEvalFix is included as an additional APR reference benchmark in Table~\ref{tab:public-bench}.

\textbf{OSS study.} An exploratory repository-level study is conducted on 10 open-source Python repositories, with 10 issues sampled per repository.

\textbf{Primary metrics.} The main reported metric is \emph{\%Resolved} for benchmark tasks, together with issue-resolution counts and summary statistics for the OSS study.

\subsection{Baselines}
The comparison includes the standard baselines, popular in coding tasks: SWE-agent~\cite{yang2024sweagent}, OpenHands~\cite{openhands2024}, AutoCodeRover~\cite{zhang2024autocoderover}, and an ablated \ours{} without RL (\ours{}\,w/o RL) and without DTG.
The numbers for the baselines are taken from the original papers and the benchmark website.~\cite{swebenchsite}

\subsection{Compute Budget and Settings}

Uniform wall-clock budget per issue (e.g., 60--120 minutes), identical tool access, and fixed seeds are used across compared systems where applicable. Summary statistics for the OSS study are reported in Table~\ref{tab:model_stats}.

\subsection{Repository-level OSS study}

In addition to public benchmarks, an exploratory repository-level study is conducted on 10 open-source Python repositories spanning low- to high-complexity maintenance settings. A total of 100 issues (10 per repository) represents a spectrum of complexity, from low-dependency utilities (jsonpickle) to high-complexity image processing libraries involving C-extensions (Pillow). The selection criteria required repositories to have active maintenance, $>1k$ stars, and a history of well-documented issue resolution to serve as the ground truth. "Complexity" is qualitatively assessed based on dependency depth and code coupling (see Table~\ref{tab:github-repos}). The purpose of this study is ecological validity; it is not intended to replace standardized benchmarks such as SWE-bench.

Each system is evaluated under the same wall-clock and tool budget. A task in each trajectory is counted as resolved only if the generated patch satisfies the issue requirements under manual inspection and all repository tests pass. Because the issue set is curated for resolvability and the context is reconstructed from the final PR, these results should be interpreted as a controlled exploratory comparison rather than as a definitive leaderboard.

\begin{table}[t]
\centering
\caption{Open-source repositories used in the OSS study (snapshot: 2025-12-05).}
\label{tab:github-repos}
\resizebox{\columnwidth}{!}{\begin{tabular}{l c c c c c}
\toprule
\textbf{Repository} & \textbf{URL} & \textbf{Stars} & \textbf{Total Issues} & \textbf{Selected Issues} & \textbf{Complexity} \\
\midrule
jsonpickle        &  \href{https://github.com/jsonpickle/jsonpickle}{https://github.com/jsonpickle/jsonpickle} & 1.3k & 62 & 10 & low\\
arrow-py & \href{https://github.com/arrow-py/arrow}{https://github.com/arrow-py/arrow} & 8.9k & 87 & 10 & low\\
flashtext & \href{https://github.com/vi3k6i5/flashtext}{https://github.com/vi3k6i5/flashtext} & 5.7k & 59 & 10 & low\\
requests & \href{https://github.com/psf/requests}{https://github.com/psf/requests} & 53.4k & 203 & 10 & low\\
poetry & \href{https://github.com/python-poetry/poetry}{https://github.com/python-poetry/poetry} & 33.9k & 518 & 10 & medium\\
Pillow & \href{https://github.com/python-pillow/Pillow}{https://github.com/python-pillow/Pillow} & 13.2k & 67 & 10 & high\\
python-qrcode & \href{https://github.com/lincolnloop/python-qrcode}{https://github.com/lincolnloop/python-qrcode} & 4.2k & 42 & 10 & low\\
tinydb & \href{https://github.com/msiemens/tinydb}{https://github.com/msiemens/tinydb} & 7.4k & 20 & 10 & medium\\
urllib3 & \href{https://github.com/urllib3/urllib3}{https://github.com/urllib3/urllib3} & 4k & 134 & 10 & medium\\
httpx & \href{https://github.com/encode/httpx}{https://github.com/encode/httpx} & 14.8k & 61 & 10 & medium\\
\bottomrule
\end{tabular}}
\vspace{-0.5em}
\end{table}

The evaluation focused on two aspects: the functional correctness of the proposed solution (checked manually) and pass/no-pass for all unit tests. A key principle of this methodology was to assess autonomous unsupervised problem resolution capabilities. 

\begin{table*}[t]
\centering
\caption{Public benchmarks for several top architectures. \%Resolved$\uparrow$ on SWE-bench and pass rate$\uparrow$ on APR benchmarks. Baseline values are from the cited papers/sites}
\label{tab:public-bench}
\begin{threeparttable}
\resizebox{\columnwidth}{!}{\begin{tabular}{lcccc}
\toprule
\textbf{Method} & \textbf{SWE-bench Verified} & \textbf{SWE-bench Lite} & \textbf{HumanEvalFix} \\
\midrule
SWE-agent + Claude 3.5 Sonnet \cite{yang2024sweagent} & 47.9 & 33.6 & -- \\
SWE-agent + Claude 4 Sonnet~\cite{yang2024sweagent} & 66.6\tnote{a} & 56.67 & 87.7\tnote{b} \\
AutoCodeRover + GPT-4o \cite{zhang2024autocoderover} & 46.2 & 37.3 & -- \\
OpenHands + Claude 3.5 Sonnet \cite{openhands2024} & 52.8 & 27.7 & -- \\
OpenHands + GPT-5 \cite{openhands2024} & 79.8 & -- & -- \\
\midrule
\textbf{AIR + Gemini 2.5 (Ours)} & \textbf{87.1} & \textbf{73.5} & -- \\
\bottomrule
\end{tabular}}
\begin{tablenotes}\footnotesize
\item[a] Reported \%Resolved on SWE-bench at time of writing~\cite{swebenchsite} 
(original data in~\cite{yang2024sweagent}).  
\item[b] HumanEvalFix pass@1 reported by SWE-agent~\cite{yang2024sweagent}; included for completeness only. \\For current SWE-bench leader-boards and variants, 
see the website~\cite{swebenchsite,swebenchverified}.
\end{tablenotes}

\end{threeparttable}
\vspace{-0.5em}
\end{table*}

\section{Ablation Studies}
\subsection{Components}
The ablation study isolates (i) RL control (\ours{}\,w/o RL), and (ii) DTG (\ours{}\,w/o Graph).
\ours{}\,w/o RL removes only the learned control policy while keeping the same DTG-based localization, editing, and validation pipeline. \ours{}\,w/o Graph removes the DTG representation and runs the same repair loop in a file-first workspace. The comparison is intended to separate the effect of repository representation from the effect of navigation/control.

\begin{table}[t]
\centering
\caption{Ablations on SWE-bench Verified (\%Resolved$\uparrow$) and operations metrics.}
\label{tab:ablations}
\begin{tabular}{l c c}
\toprule
\textbf{Variant} & \textbf{\%Resolved$\uparrow$} & \textbf{MTTR (min)$\downarrow$} \\
\midrule
\ours{} (full)         &\textbf{87.1} & \textbf{8.9} \\
\ours{}\,w/o RL        &  82.5 & 1.5 \\
\ours{}\,w/o Graph     &  56.2 & 4.6\\
\bottomrule
\end{tabular}
\vspace{-0.5em}
\end{table}

\section{Results and Analysis}
\label{sec:results}

\subsection{Ablations: What matters and why}

Table~\ref{tab:ablations} isolates the contributions of the two main design choices. Removing RL control reduces \%Resolved from 87.1 to 82.5. Although MTTR becomes shorter, this is not a gain in efficiency: the agent terminates earlier and explores fewer candidate repair trajectories. Removing DTG causes a much larger drop, to 56.2, indicating that repository representation is the dominant factor for multi-file fault localization. In other words, RL improves search control, but DTG provides the causal structure that makes the search tractable in the first place.

Overall, \textsc{RL} and \textsc{Graph} are the largest drivers of the resolution rate. \label{sec:ablation-analysis}

\subsection{Comparison with commercial tools}

Table~\ref{tab:github-bench} and Table~\ref{tab:model_stats} summarize model performance across 10 OSS projects (100 total issues). This study is less rigorous than SWE-bench for two reasons. First, the sample size is smaller. Second, 10 issues per repository were pre-selected for likely resolvability, which introduces selection bias toward coding issues that can be addressed through standard debugging and repair practice.

For each case, the evaluation context was reconstructed from the final pull request that resolved the issue.

A score of +1 is assigned when the issue requirements are satisfied and all repository unit tests pass. No penalty is assigned for time, cost, or failed attempts. Figure~\ref{fig:perf} shows the per-project comparison; the full results are reported in Table~\ref{tab:github-bench}, with summary statistics in Table~\ref{tab:model_stats}.

\begin{figure*}[t]
\centering
\begin{tikzpicture}
\begin{axis}[
    ybar,
    bar width=12pt,
    width=\textwidth,
    height=7cm,
    ymin=0, ymax=10,
    enlarge x limits=0.12,
    ylabel={Issues resolved (\#)},
    xlabel={},
    symbolic x coords={jsonpickle,arrow-py,flashtext,requests,poetry},
    xtick=data,
    legend style={at={(0.5,1.05)}, anchor=south, legend columns=4, row sep=0.5pt},
    nodes near coords,
    nodes near coords align={vertical},
]
\addplot coordinates {(jsonpickle,5) (arrow-py,4) (flashtext,2) (requests,3) (poetry,1)};
\addplot coordinates {(jsonpickle,6) (arrow-py,4) (flashtext,2) (requests,8) (poetry,3)};
\addplot coordinates {(jsonpickle,9) (arrow-py,6) (flashtext,8) (requests,8) (poetry,4)};

\legend{Claude 4 Sonnet, Gemini-2.5 pro, \ours{} (this work)}
\end{axis}
\end{tikzpicture}
\caption{Issue-resolution results on top 5 OSS repositories (10 issues per repository).}
\label{fig:perf}
\end{figure*}

\begin{table*}[t]
\centering
\caption{Issue resolution on ten OSS GitHub repos (10 issues each; same wall-clock + tool budget). Numbers indicate how many issues were resolved.}
\label{tab:github-bench}
\begin{threeparttable}
\begin{tabular}{lccccc}
\toprule
\textbf{Project} & \textbf{\ours{} (this work)} & \textbf{Claude 4 Sonnet} & \textbf{Gemini-2.5 pro}\\
\midrule
jsonpickle & \textbf{9} & 5 & 6 \\
arrow-py & \textbf{6} & 4 & 4 \\ 
flashtext & \textbf{8} & 2 & 2 \\
requests & \textbf{8} & 3 & 8 \\
poetry & \textbf{4} & 1 & 3 \\
Pillow & \textbf{4} & 0 & 0 \\
python-qrcode & \textbf{8} & 1 & 2 \\
tinydb & 5 & \textbf{6} & 6 \\
urllib3 & \textbf{9} & 2 & 7 \\
httpx & \textbf{9} & 4 & 3 \\
\bottomrule
\end{tabular}
\vspace{-0.5em}
\end{threeparttable}
\end{table*}

\begin{table}[h]
    \centering
    \caption{Performance statistics per model across 10 projects (100 total issues).}
    \label{tab:model_stats}
    \begin{tabular}{lrrr}
        \toprule
        \textbf{Statistic} & \textbf{AIR (This Work)} & \textbf{Claude 4 Sonnet} & \textbf{Gemini-2.5 pro} \\
        \midrule
        Total Solved & 70 & 28 & 41 \\
        Success Rate & 70.0\% & 28.0\% & 41.0\% \\
        Mean (per project) & 7.0 & 2.8 & 4.1 \\
        Median & 8.0 & 2.5 & 3.5 \\
        Std. Dev. & 2.05 & 1.93 & 2.56 \\
        Min / Max & 4 / 9 & 0 / 6 & 0 / 8 \\
        \bottomrule
    \end{tabular}
\end{table}

Key Observations:

\begin{itemize}
    \item AIR achieved the highest score among the compared systems, solving 70\% of all issues. It was the only model to solve at least 4 issues in every project.
    
    \item Gemini-2.5 pro came in second with 41\%, showing strong performance on specific libraries like requests (8/10) and urllib3 (7/10) but struggling with Pillow (0/10).

    \item Claude 4 Sonnet solved 28\% of issues. It had a high variance, performing well on tinydb (6/10) but solving 2 or fewer issues in half of the projects.

    \item Difficulty: Pillow appeared to be the hardest project, with the two commercial models scoring 0 points and AIR scoring its minimum of 4 points. jsonpickle and requests generally saw higher success rates across the board.

\end{itemize}

\subsubsection{Performance Analysis and the "Semantic Gap"}

The results in Figure~\ref{fig:perf} and Table~\ref{tab:github-bench} reveal a strong correlation between the explicitness of the data flow and the agent's success rate. AIR achieves a 70\% resolution rate overall, showing particular strength in "loose-coupling" architectures like requests and urllib3 (8/10 and 9/10 respectively). In these repositories, logic errors are often distributed across multiple abstraction layers - a scenario where the DTG's ability to trace data lineage allows the agent to "hop" over irrelevant middleware, a capability lacking in the file-based baselines.

\subsubsection{Failure Case Study: The Pillow Repository}

A notable deviation is observed in the Pillow repository, where commercial models scored 0/10 and AIR achieved only 4/10. This performance drop stems from the "Language Boundary" problem. Pillow relies heavily on C-extensions for image processing. The current DTG implementation parses Python ASTs but treats C-bindings as opaque edges. Consequently, when data flows into a compiled extension, the lineage is broken, blinding the agent. This limitation highlights the necessity for future work in cross-language graph construction to support hybrid-runtime environments.

\section{Limitations}

The present results should be interpreted with three limitations. First, the current DTG builder primarily targets Python and treats cross-language boundaries (e.g., C/C++ extensions) conservatively, which reduces observability in mixed-language repositories. Second, the OSS study uses a curated issue set and reconstructed context, so it is less rigorous than standardized benchmarks and may overestimate deployment-time performance. Third, AIR is evaluated empirically; the current version does not provide formal correctness guarantees beyond repository tests and static/dynamic validation signals. Consequently, AIR should be viewed as a practical empirical repair system rather than a formally verified maintenance framework.

\section{Conclusion and Future Work} \label{sec:conclusion}

This paper has argued that the dominant file-first, control-centric view of repositories is misaligned with both human debugging practice and the causal structure of many real bugs. By re-centering our representation around data states and transformations, the \emph{Data-First Transformation Graph} (DTG) offers a compact and causally meaningful substrate for LLM-based repair agents.

The empirical results on SWE-bench Verified, where AIR achieved an 87.1\% resolution rate, validate that shifting from text-based retrieval to graph-based navigation substantially reduces the search complexity of logic repair. However, our analysis of failure cases, particularly within the Pillow repository, highlights critical avenues for future development. We outline three primary directions for the evolution of AIR:

\textbf{1. Breaking the Language Boundary (Polyglot DTGs):} The experiments revealed that the current DTG construction, while effective for pure Python, struggles with hybrid runtimes (e.g., Python wrappers around C/C++ extensions). Future iterations must implement cross-language graph builders that can trace the data lineage through Foreign Function Interfaces (FFI). This involves unifying AST parsers across languages into a shared intermediate representation, allowing the agent to "tunnel" through compiled extensions rather than treating them as opaque edges.

\textbf{2. Dynamic Graph Refinement via RL:} Currently, the DTG is static, constructed prior to the agent's traversal. This paper proposes a dynamic architecture where the agent can modify the graph structure at runtime—collapsing well-tested subgraphs to save context or expanding ambiguous nodes for deeper inspection. By integrating this into the Reinforcement Learning policy, the agent can learn to optimize its own environment representation, effectively performing "attention management" over the repository topology.

\textbf{3. Beyond Repair: Security and Refactoring:} The utility of the DTG extends beyond Automated Program Repair. We envision applying this data-centric view to automated security auditing, where "taint analysis" becomes a native graph traversal task rather than a complex heuristic. Similarly, the DTG offers a robust foundation for architectural refactoring, allowing agents to identify and decouple "god objects" based on data cohesion metrics rather than just line counts.

Overall, AIR suggests that data-lineage-aware navigation can improve repository -scale repair when compared to file-first interaction alone. The main contribution of this work is therefore not a claim of universal autonomy but a concrete data-first representation and agent design that improves causal localization under repository-scale context constraints. The results are encouraging, while the identified limitations define the work still required for a larger deployment.

\subsubsection{Acknowledgments.} The author thanks Modal and lightning.ai for providing compute resources and free cloud credits for our experiments. 

\section*{Appendix 1}

\begin{lstlisting}[language=Python, caption=A code snippet from poetry project - some non-essential lines were cut for brevity (poetry/ src/poetry/installation/chooser.py:125:197)]
    def _no_links_found_error(
        self,
        package: Package,
        links_seen: int,
        wheels_skipped: int,
        sdists_skipped: int,
        unsupported_wheels: set[str],
    ) -> PoetryRuntimeError:
        messages = []
        info = (
            f"This is likely not a Poetry issue.\n\n"
            f"  - {links_seen} candidate(s) were identified for the package\n"
        )

        messages.append(ConsoleMessage(info.strip()))

        if unsupported_wheels:
            messages += [
                ConsoleMessage(
                    "The following wheel(s) were skipped as the current project environment does not support them "
                    "due to abi compatibility issues.",
                    debug=True,
                ),
                ConsoleMessage("\n".join(unsupported_wheels), debug=True)
                .indent("  - ")
                .wrap("warning"),
                ConsoleMessage(
                    "If you would like to see the supported tags in your project environment, you can execute "
                    "the following command:\n\n"
                    "    <c1>poetry debug tags</>",
                    debug=True,
                ),
            ]

        source_hint = ""
        if package.source_type and package.source_reference:
            source_hint += f" ({package.source_reference})"

        return PoetryRuntimeError(
            reason=f"Unable to find installation candidates for {package}",
            messages=messages,
        )
\end{lstlisting}

\begin{figure}[!t]
\centering
\includegraphics[width=4.8in]{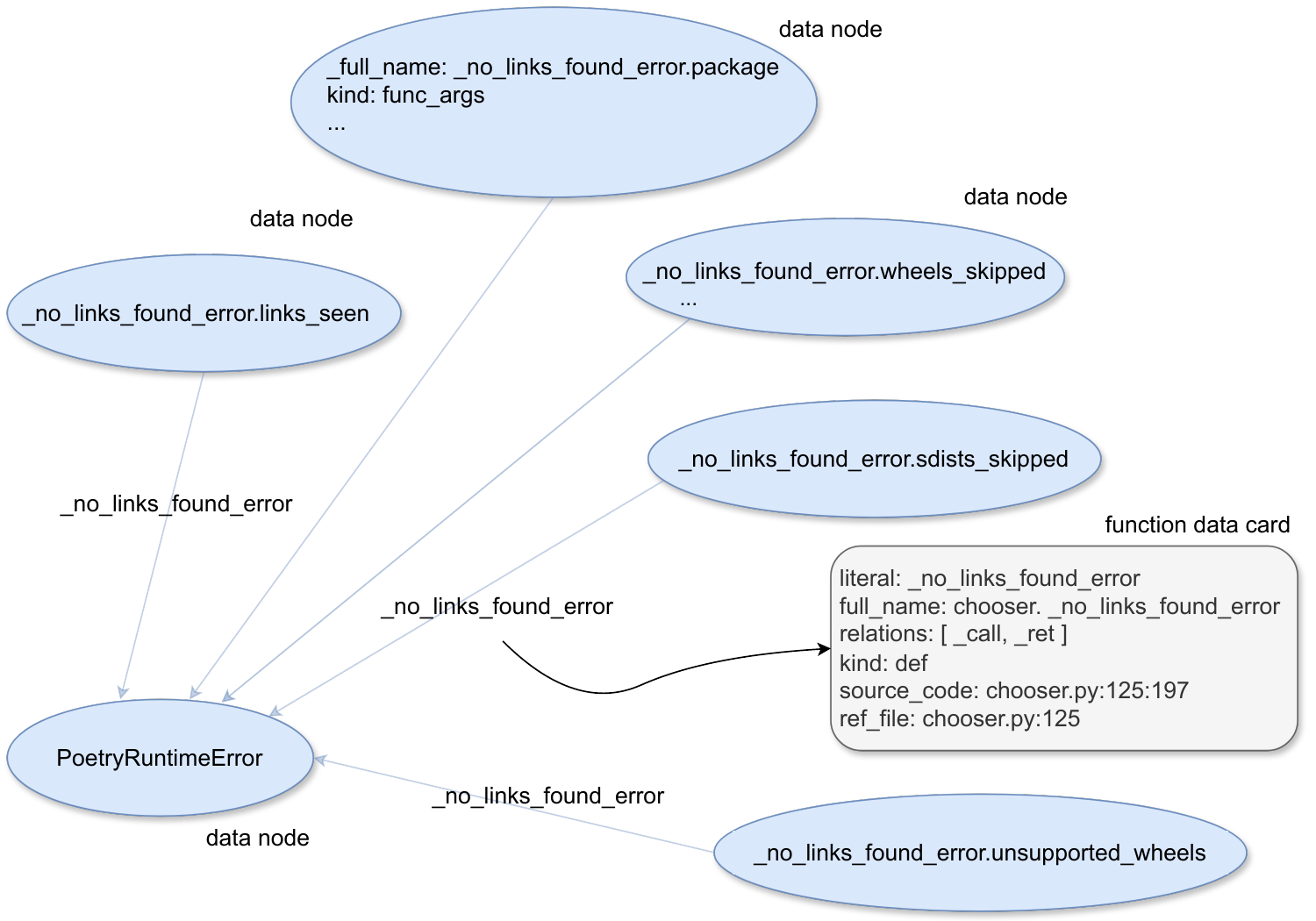}
\caption{The reconstructed sample of Data Transformation Graph given the raw data to build Nodes and Edges from Table~\ref{tab:constructs}.}
\label{fig:sample_dtg}
\end{figure}

The graph builder can extract the following raw constructs from this snippet of code (Table~\ref{tab:constructs}) and build the subgraph depicted in Fig.~\ref{fig:sample_dtg}. The standard Python \texttt{ast} library is used for parsing Python files, Neo4j for storing nodes and edges, and ChromaDB as a vector-storage tier for semantic search over relevant nodes and edges.

\begin{table}[h]
    \centering
    \begin{tabular}{l c l}
        \hline
        \textbf{Construct Category} & \textbf{Count} & \textbf{Items identified} \\
        \hline
        Function Definitions & 1 & \texttt{\_no\_links\_found\_error} \\
        Arguments & 6 & \texttt{self}, \texttt{package}, \texttt{links\_seen}, \texttt{wheels\_skipped}, \\
                  &   & \texttt{sdists\_skipped}, \texttt{unsupported\_wheels} \\
        Local Variable Assignments & 3 & \texttt{messages}, \texttt{info}, \texttt{source\_hint} \\
        Control Flow (Branches) & 4 & \texttt{if} statements \\
        Return Statements & 1 & Returns \texttt{PoetryRuntimeError} \\
        External/Global References & 3 & \texttt{Package}, \texttt{PoetryRuntimeError}, \texttt{ConsoleMessage} \\
        \hline
    \end{tabular}
    \vspace{0.1in}
    \caption{Summary of extracted code constructs. Verbose metadata such as file name, line number, and variable type is omitted for brevity. Function arguments, return values, and local variables become data nodes. The function name forms the basis for the corresponding transformation edge. The resulting structures are stored in Neo4j, while code embeddings are stored in ChromaDB.}
    \label{tab:constructs}
\end{table}

\newpage

\section*{Appendix 2}

Table~\ref{tab:landscape} presents a comparison of the market tier, interaction modality, and primary capabilities of current AI coding tools. AIR is positioned differently in its focus on asynchronous, autonomous maintenance.

\begin{table*}[t]
\centering
\small
\caption{Representative AI-assisted software tools (Q4 2025). The comparison distinguishes market tier, interaction modality, primary capability, and operational scope relevant to software maintenance automation.}
\label{tab:landscape}
\begin{tabular}{@{}p{2.1cm}p{1.7cm}p{1.8cm}p{2.2cm}p{4.3cm}@{}}
\toprule
\textbf{System} & \textbf{Tier} & \textbf{Modality} & \textbf{Primary capability} & \textbf{Scope / distinguishing characteristic} \\
\midrule
GitHub Copilot & Professional & IDE/editor plugin & Code completion, chat, refactoring & Strong ecosystem integration; optimized for interactive developer assistance. \\
Amazon CodeWhisperer & Professional & IDE plugin & Code completion, security hints & Security-oriented suggestions integrated with cloud development workflows. \\
Tabnine & Professional & IDE plugin & Local/cloud completion & Emphasis on privacy-preserving deployment and enterprise control. \\
Cursor & Professional & AI-first IDE & Conversational editing, repository search & Editor centered on agentic edits with strong repository-context support. \\
Claude Code & Professional & Chat + tools & Conversational code reasoning & Long-context code reasoning with strong natural-language interaction. \\
OpenAI Codex (legacy) & Professional & API/chat (retired) & Code generation foundation model & Historical foundation system that influenced later tool development. \\
Sourcegraph & Professional & IDE/plugin platform & Repository reasoning, code editing & Broad repository navigation and assistant functionality across large codebases. \\
\addlinespace
B44 (builder) & Basic & Web builder & Prompt-based app scaffolding & Rapid MVP creation; limited support for repository-scale maintenance. \\
Lovable & Basic & Web builder & No/low-code generation & Accessible rapid prototyping; limited support for multi-file change management. \\
Replit (Ghostwriter) & Basic & Web IDE & In-editor assistance & Lightweight assistance oriented toward education and small-team workflows. \\
\addlinespace
AlphaCode~\cite{li2022alphacode} & Specialized & Research system & Competition-style code generation & Strong performance on competitive programming tasks; not designed for repository maintenance. \\
AlphaEvolve~\cite{deepmind2025alphaevolve} & Specialized & Research system & Algorithm discovery/optimization & Effective in narrow optimization settings; not a general software maintenance system. \\
Diffblue Cover & Specialized & JVM toolchain & Unit test synthesis & Industrial-grade test generation; complementary to, rather than a substitute for, issue repair. \\
Getafix/Infer lineage~\cite{bader2019getafix} & Specialized & Internal/OSS tools & Learned fix patterns, static analysis & Pattern-based repair pipeline requiring suitable bug/fix corpora and analysis infrastructure. \\
Factory.ai & Specialized & Cloud integration & Agent-native software development & Enterprise-oriented maintenance workflow support. \\
\bottomrule
\end{tabular}
\end{table*}

\clearpage
%
%
\bibliographystyle{unsrt}
\bibliography{reference}
\end{document}